\documentclass[10pt,journal,final,twocolumn]{IEEEtran}
\usepackage{amssymb}
\usepackage{graphicx}
\usepackage{color}
\usepackage{times}
\usepackage{epsfig}
\usepackage{amsmath}
\usepackage{amssymb}
\usepackage{caption}
\usepackage{subfig}
\usepackage{multirow}
\usepackage{lscape}
\usepackage{rotating}
\usepackage{url}
\usepackage{array}

\allowdisplaybreaks

\input{Definitions}
\graphicspath{{./}{./Figures/}}

\begin{document}

\title{Synthesizing Filamentary Structured Images with GANs}

\author{\IEEEauthorblockN{He Zhao\IEEEauthorrefmark{1}\IEEEauthorrefmark{2},
Huiqi Li\IEEEauthorrefmark{1},
Li Cheng\IEEEauthorrefmark{2}\IEEEauthorrefmark{3}} \\
\IEEEauthorblockA{\IEEEauthorrefmark{1}Beijing Institute of Technology, China} \\
\IEEEauthorblockA{\IEEEauthorrefmark{2}Bioinformatics Institute, A*STAR, Singapore} \\
\IEEEauthorblockA{\IEEEauthorrefmark{3}School of Computing, National University of Singapore, Singapore}
\thanks{
Huiqi Li and Li Cheng are the correspondence authors.}}

\maketitle

\begin{abstract}
This paper aims at synthesizing filamentary structured images such as retinal fundus images and neuronal images, as follows: Given a ground-truth, to generate multiple realistic looking phantoms.
A ground-truth could be a binary segmentation map containing the filamentary structured morphology, while the synthesized output image is of the same size as the ground-truth and has similar visual appearance to what have been presented in the training set.
Our approach is inspired by the recent progresses in generative adversarial nets (GANs) as well as image style transfer.
In particular, it is dedicated to our problem context with the following properties:
Rather than large-scale dataset, it works well in the presence of as few as 10 training examples, which is common in medical image analysis; It is capable of synthesizing diverse images from the same ground-truth; Last and importantly, the synthetic images produced by our approach are demonstrated to be useful in boosting image analysis performance.
Empirical examination over various benchmarks of fundus and neuronal images demonstrate the advantages of the proposed approach.
\end{abstract}

\section{Introduction}

A rather broad range of important biomedical images are filamentary structured in nature, including retinal fundus images, neuronal images, 
among many others.
Take fundus images for example, topological and geometrical properties of vessel filamentary structures provide valuable clinical information in diagnosing diseases such as proliferative diabetic retinopathy, glaucoma, and hypertensive retinopathy~\cite{AbrGarSon:tmi10}. Often, only a handful of annotated images are available, where the filamentary structures are delineated by domain experts through laboriously manual process --- a typical situation that exists in many biomedical applications.

On the other hand, image analysis and image synthesis have long been regarded as tightly intertwined techniques, which include e.g. the research works in analysis-by-synthesis~\cite{Gre:book76-81},
as well as the recent progresses in human full-body and hand pose estimation problems~\cite{ShoEtAl:pami13,XuChe:iccv13}, where the synthesis processes or synthesized data play an important role in addressing the analysis tasks.
In the meantime, although a large number of research activities have been engaged in filamentary structured image analysis (see e.g. the survey papers of retinal and neuronal image analyses~\cite{AbrGarSon:tmi10,FrazEtAl:CMPB12,PenEtAl:neuron15}), relatively very little effort exists concerning synthesizing such images.

In this paper, we present a learning based approach for synthesizing filamentary structured biomedical images. Our paper possesses the following major contributions:
First, our synthesis model can be effectively learned in data-driven fashion from a relatively small sample size. For example, we have successfully constructed synthesis models for STARE~\cite{HooEtAl:tmi00} and DRIVE~\cite{StaEtAl:tmi04} fundus image benchmarks, where the corresponding training images are merely 10 and 20 images, respectively.
Second, based on a single ground-truth input, our approach is capable of synthesizing multiple distinct images. This capacity of introducing diversity is important in biomedical data simulations.
Third, the synthesized images are shown useful in improving image segmentation performance. In other words, a baseline supervised segmentation module is shown to improve its segmentation performance on real-world datasets when trained with additional images synthesized by our approach.
Fourth, to our knowledge our approach\footnote{Note our implementation, together with the related results, are made publicly available at \url{https://web.bii.a-star.edu.sg/archive/machine_learning/Projects/filaStructObjs/Synthesis/index.html}.} is among the first attempt toward data-driven synthesis of filamentary structured images, including fundus and neuronal images. More specifically, two variants of our proposed pipeline have been proposed and studied, which we refer to as \emph{Fila-GAN} and \emph{Fila-sGAN}, respectively.
Extensive experiments on various applications and datasets demonstrate their ability in producing realistic looking images, as well as in boosting the performance of a baseline segmentation module.

\section{Related Works}


\paragraph*{Synthesizing filamentary structured images}
There have been several existing research activities in simulating filamentary structured images. 
%
%
In synthesizing retinal vascular structures, one notable application area is surgical simulations~\cite{SagEtAl:cgit94}.
Others are driven by the practical demand in empirical evaluations of segmentation or tracing methods.
A critical issue with retinal image analysis lies in the lack of ground-truth annotations, due to its expensive and laborious nature.
This is further complicated by the inter- and intra- observer variabilities of expert human observers that are subjective and prone to annotation errors~\cite{friEtal:bookchapter02}.
Synthesizing retinal phantoms~\cite{FioEtAl:stag14,MentiEtAl:SASHIMI16} may be useful in this context given its unique advantage of having complete and unambiguous knowledge of the corresponding ground-truths.
In~\cite{FioEtAl:stag14}, the specific vascular morphology and texture are constructed from scratch based heavily on domain knowledge.
The work of~\cite{MentiEtAl:SASHIMI16} instead aims to derive vessels and textures from real data, which is based on their morphological and textural statistics and by utilizing active shape contour and Kalman filter techniques.
Neuronal image synthesis has also been studied with significant biological prior knowledge, where GENESIS, NEURON, and L-Neuron are probably the most well-known efforts.
GENESIS~\cite{BowCorBee:bookchapter14} is a simulation environment for constructing realistic models of neurobiological systems. It was one of the first simulation systems specifically designed for modeling nervous systems;
%
NEURON~\cite{CarHin:book06} is similarly developed as a simulation environment for modeling individual neurons and networks of neurons;
L-Neuron~\cite{AscKri:neurocomp00} generates and studies anatomically accurate neuronal phantoms based on sets of recursive rules described by a Lindenmayer system or L-system.
Different from the past efforts, our goal in this paper is to synthesize realistic-looking retinal phantoms in an end-to-end fashion with data-driven approach. 

\paragraph*{Image style transfer, filamentary structured image segmentation, and data augmentation}
The problem of image style transfer has been studied in the past twenty years~\cite{HerEtAl:SIGGRAPH01,CheVisZha:CVPR08}. Recently impressive results are obtained by Gatys et al.~\cite{GatEckBet:arxiv15} through the successful application of deep learning techniques. It has been further improved by a number of works including~\cite{UlyEtAl:icml16,JusEtAl:ECCV16} with a preference toward being efficient and light-weight.
To our knowledge, the proposed approach is the first such attempt to incorporate style transfer techniques in simulating filamentary structured images.
Meanwhile, there has been vast literature on filamentary structured image segmentation,
interested readers may refer to~\cite{KirQue:acmcs00,LesEtAl:MidIA09} for more thorough reviews.
In spite of these research efforts, it remains challenging to precisely segment 2D and 3D image-based filaments.
This is evidenced by e.g. the recent BigNeuron initiative~\cite{PengEtAl:neuroinfo15} that calls for innovations in addressing
the demands from neuronal science community where a significant number of neuronal images have been routinely produced in wet labs,
while there still lack sufficiently accurate tools to automatically segment the neurite structures.
Our work may shed lights on the feasibility of utilizing the generated phantoms in improving segmentation performance.
Furthermore, there have been research efforts to enrich the training dataset by means of data augmentation, such as cropping, flipping and rotating existing training images~\cite{szegedy2015going, ciregan2012multi} as well as applying small perturbations in the color or intensity imaging space~\cite{krizhevsky2012imagenet}.
In a sense, contrary to the aforementioned methods that mechanically memorizes the original image style or texture, our approach could be regarded as a more systematic way to enrich existing training examples with phantoms of similar style.

\paragraph*{Generative adversarial networks}
The advancement of deep learning techniques~\cite{KinWel:ICLR14,GooEtAl:NIPS14,OorKalKav:ICML16} have led to significant progress in generating photo-realistic images using techniques such as generative adversarial networks (GANs)~\cite{GooEtAl:NIPS14}. 
It considers a two-player zero-sum game between two agents, a discriminator net and a generator net. The discriminator is to tell the real inputs apart from the faked ones, while the role of the generator is to fool the discriminator by synthesizing instances that appear to have come from the real data distribution.
%
Subsequently, a number of GANs variants~\cite{RadMetChi:arxiv15,MirOsi:arxiv14,denton2015lapgan,chen2016infogan,arjovsky2017wasserstein} have been developed. 
Among them, DCGAN~\cite{RadMetChi:arxiv15} introduces a set of constraints to stabilize the training dynamics between the generator and the discriminator.
CGAN~\cite{MirOsi:arxiv14} facilitates the training of a synthesis model to generate images conditioned on some auxiliary information.
LAPGAN~\cite{denton2015lapgan} uses a cascade of convolutional networks within a Laplacian pyramid framework to generate the images from coarse to fine.
InfoGAN~\cite{chen2016infogan} allows to learn disentangled representations in a completely unsupervised manner.

The closest work is perhaps that of~\cite{IsoEtAl:arxiv16}, which also utilizes techniques similar to U-Net~\cite{RonFisBro:miccai15} 
to preserve global structural information during data generation process. 
Different from ~\cite{IsoEtAl:arxiv16} which tends to generate only fixed output for a given input, our approach is able to produce unlimited number of individual phantoms from the same input. Moreover, instead of working with few hundreds to millions of training examples as of~\cite{IsoEtAl:arxiv16}, our approach works with only tens of training images. Finally, our results are shown to boost the retinal segmentation performance when being included as additional training examples.

\begin{figure*}
\centering
\scalebox{0.82}{
\includegraphics[width=0.99\textwidth]{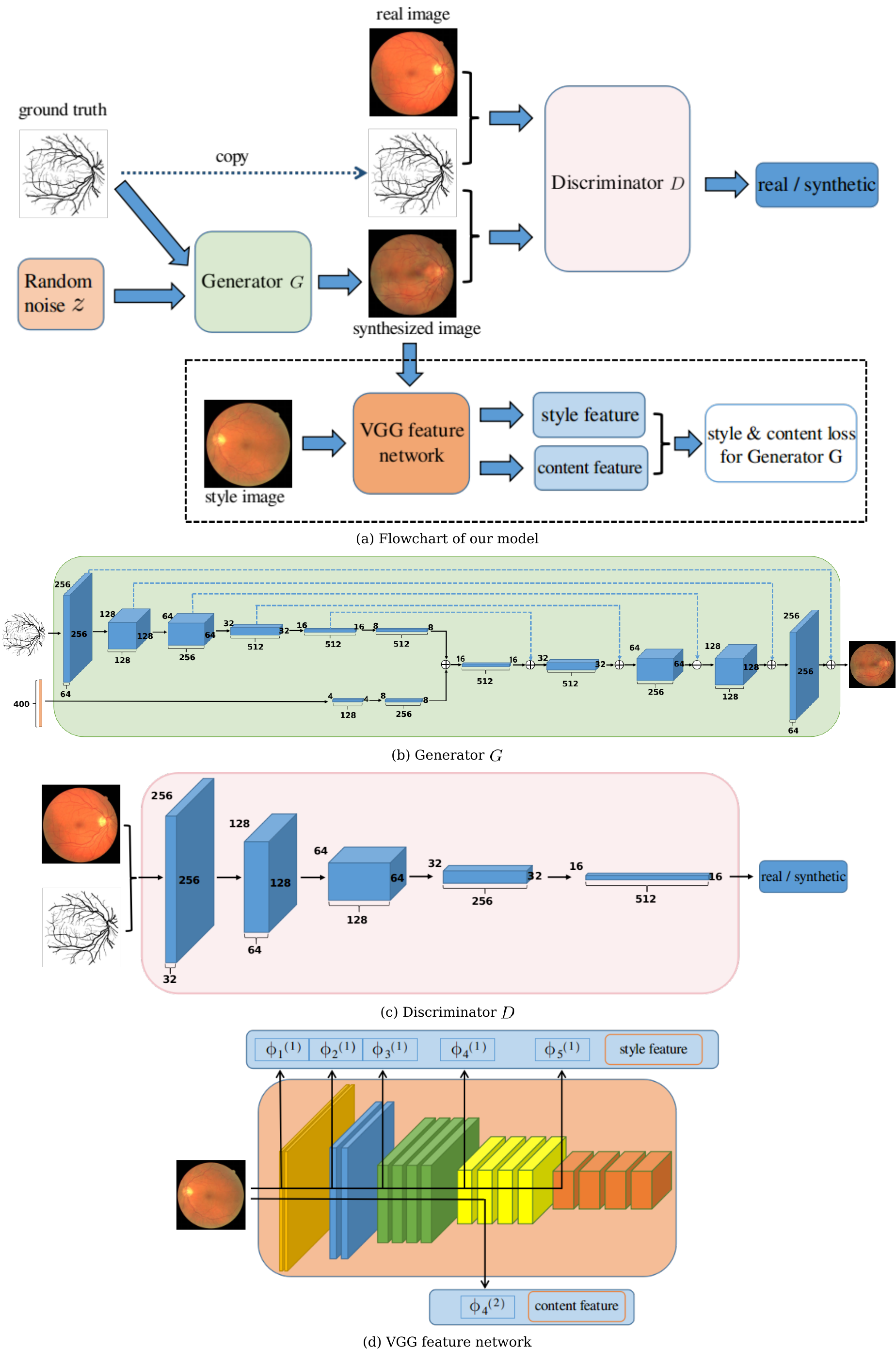}
}
\caption{\footnotesize{(a) Flowchart of our approach, which contains the generator and the discriminator networks as detailed in (b) and (c). The dimensions of all layers are specified. The VGG feature networks are described in (d), where the top row indicates the specific layers (e.g. $\phi_{1}^{(1)}$, $\phi_{2}^{(1)}$, $\ldots$) extracted as style features, while bottom row are the layers (e.g. $\phi_{4}^{(2)}$) used as content features. See text for details.}}
\label{fig:flowchart}

\end{figure*}

\section{Our Approach}

In this work, we aim to learn a direct mapping from a segmentation (could be a ground-truth) 
back to a plausible raw filamentary structured image.
More specifically, denote $\vec{x} \in \mathbb{R}^{W\times H\times 3}$ a RGB filamentary structured image, $\vec{y} \in \{0, 1\}^{W\times H}$ the corresponding segmentation (i.e. ground-truth annotation). By imitating the image formation processing, let $G_{\theta}: \left( \vec{y} \in \mathbb{R}^{W\times H}, \vec{z} \in \mathbb{R}^{Z} \right) \rightarrow \hat{\vec{x}} \in \mathbb{R}^{W\times H\times 3}$ denote the image generation function that takes a ground-truth binary image $\vec{y}$ and a noise code $\vec{z}$ as input, to produce a filamentary structured phantom $\hat{\vec{x}}$.
Our goals are four-fold: (1) Learn the $\theta$-parameterized function $G$ from a usually very small training set $\left\{ \left(\vec{x}_i, \vec{y}_i \right) \right\}_{i=1}^n$; (2) Be capable of exploring the underlying conditional image formation distribution $p(\vec{X}|\vec{y})$, where $\vec{X}$ is a random variable denoting the feasible filamentary structured image realization conditioning on the particular realization $y$. In other words, to sample plausible yet distinct RGB image instances from a same input $\vec{y}$, obtained by varying the noise code $\vec{z}$; (3) These synthesized images are useful in boosting the performance of a supervised segmentation method when adding up to the training set;
(4) Consider an interesting variant of our approach where a specific image style obtained from \emph{one} additional input image $\vec{x}_{\mathrm{s}}$ can be directly transferred to the output phantom $\vec{\hat{x}}$. Here the style of $\vec{x}_{\mathrm{s}}$ could be very different from that of $\vec{x}$, and the corresponding contents (i.e segmentations $\vec{y}_{\mathrm{s}}$ and $\vec{y}$) are usually unrelated.
The aforementioned goals seems daunting: Given the intricate nature of the image formation process,
$G$ could be a rather sophisticated function, and the situation is exacerbated by the small $n$ nature of the problem. 
Nonetheless, by resorting to the powerful deep learning framework of GANs, an end-to-end learning machine is proposed, as depicted in Fig.~\ref{fig:flowchart}.

In addition to the generator $G_{\theta}$, consider a discriminator function $D_{\gamma}:  \left( \vec{X} \in \mathbb{R}^{W\times H\times 3}, \vec{y} \in \mathbb{R}^{W\times H} \right) \rightarrow d \in [0, 1]$, whose role is to tell apart synthesized phantom $\vec{X}:=\vec{\hat{x}}$ (ideally $d \rightarrow 0$) from real filamentary structured image $\vec{X}:=\vec{x}$ (ideally $d \rightarrow 1$), as visually explained in Fig.~\ref{fig:flowchart}(a). We follow the GANs idea for this two-player zero-sum game setting and consider the following optimization problem that characterizes the interplay between $G$ and $D$:
\begin{align}
\label{eq:minmax}
&\min_{\theta}\max_{\gamma} L \left(G_{\theta}, D_{\gamma}\right)
=  \mathbb{E}_{\vec{x}, \vec{y} \sim p(\vec{x},\vec{y})} \left[\log D_{\gamma}(\vec{x}, \vec{y}) \right] + \\
\nonumber
&\mathbb{E}_{\vec{y}\sim p(\vec{y}), \vec{z}\sim p(\vec{z})} \left[\log \left(1 - D_{\gamma} \left(G_{\theta}(\vec{y}, \vec{z}), \vec{y} \right)\right)\right]
+ \lambda L_{\mathrm{DEV}}(G_{\theta}),
\end{align}
with $\lambda>0$ being a trade-off constant.
Here the last term is introduced to ensure the synthesized image will not deviate significantly from the real image, and we consider a simple L1 loss
\begin{align}
\label{eq:dev_loss}
L_{\mathrm{DEV}}(G_{\theta}) = \mathbb{E}_{\vec{x}, \vec{y} \sim p(\vec{x},\vec{y})} \left[\left\| \vec{x} - G_{\theta}(\vec{y}, \vec{z}) \right\|_1 \right].
\end{align}
In summary, during training generator $G$ tries to synthesize realistic-looking images that can fool the discriminator $D$, by minimizing the objective function of Eq.~\eqref{eq:minmax}. In practice, following the approximation scheme of~\cite{GooEtAl:NIPS14}, it is realized by minimizing a simpler form $-\log(D_{\gamma}(G_{\theta}(\vec{y}, \vec{z})))$, instead of the original $\log(1-D_{\gamma}(G_{\theta}(\vec{y}, \vec{z})))$. To summarize, learning the generator $G$ amounts to minimizing
{\small
\begin{align}
\label{eq:g_loss}
L_{\mathcal{G}}(G_{\theta}) = - \sum_i \log D_{\gamma} \left(G_{\theta} \left(\vec{y}_i,\vec{z}_i \right), \vec{y}_i \right) + \lambda \left\|\vec{x}_i- G_{\theta} \left(\vec{y}_i,\vec{z}_i \right) \right\|_1.
\end{align}
}
On the other hand, discriminator $D$ attempts to properly separate the real images from synthesized ones by maximizing Eq.~\eqref{eq:minmax}.
while discriminator $D$ is learned by maximizing
\begin{align}
\label{eq:d_loss}
L_{\mathcal{D}}(D_{\gamma}) =  \sum_i \log D_{\gamma}(\vec{x}_i,\vec{y}_i) + \log\left( 1-D_{\gamma} \left(G_{\theta} \left(\vec{y}_i,\vec{z}_i \right), \vec{y}_i \right) \right).
\end{align}
Note the empirical sum is used in practice to approximate the expectation.
The learning process is practically carried out by alternating between these two optimization operations that are similarly adopted by GANs and variants~\cite{GooEtAl:NIPS14,RadMetChi:arxiv15,IsoEtAl:arxiv16}. Unfortunately there does not exist a formal guarantee that this optimization procedure will reach Nash equilibrium. That being said, in practice we have observed convergence traits with reasonable image synthesis outputs, as also been demonstrated in the above GANs related works.
The top part of Fig.~\ref{fig:flowchart}(a) (i.e. excluding elements in the dashed box) illustrates the overall work flow of our approach so far, also referred to as \emph{Fila-GAN}. In the following, we describe in detail the specific neural net architectures of our functions $G$ and $D$.

\subsection{Generator $G$ and Discriminator $D$}
The commonly used encoder-decoder strategy~\cite{WanGup:ECCV16,MaoSheYan:NIPS16,PatEtAl:CVPR16,IsoEtAl:arxiv16}
is adopted here, which allows the introduction of noise code in a natural manner. The encoder part acts as a feature extractor where the multiple layered structure captures local to more global data representations.
The $400$-dimensional random code $\vec{z}$ is fully connected to the first layer, which is then reshaped.
Note that unless otherwise specified, for all layers of $G$ and $D$, we use kernel size 4 and stride 2 without any pooling layer.
Meanwhile, in our context it is crucial for the generator function to respect the input filamentary structured morphology during image generating process. As such, the skip connections of U-Net~\cite{RonFisBro:miccai15} are also considered here. That is, previous layer is duplicated by appending onto current layer in a mirrored fashion that skips odd-number of layers with the center coding layer as its origin.
It is worth noting that when image size is small and the network model is shallow, the encoder-decoder framework may work well even without any skip connection. However, as we are working with large image size (e.g. $512 \times 512$) and deeper networks, training such model becomes problematic. This might be due to the effects of vanishing gradients over long paths in error back-propagation. The skip connection, by playing a similar role to the residual nets~\cite{HeEtAl:CVPR16}, allows the additional direct flow of error gradients from decoder layers to corresponding encoder layers.
This facilitates the proper memorization of global and local shape contents as well as the corresponding textures encountered in the training set, and empirically helps to generate much better results.
We follow the basic network architecture proposed in~\cite{RadMetChi:arxiv15}, to build the layers of the generator with multiple Convolution-BatchNorm-LeakyRelu components as in Fig.~\ref{fig:flowchart}(b). The activation function of the output layer is $tanh$ to squash the value between -1 and 1.
On par with our generator, the same Convolution-BatchNorm-LeakyRelu building blocks are used in building our discriminator as in Fig.~\ref{fig:flowchart}(c). Here the activation function of the output layer is instead $sigmoid$.
The feature map sizes are halved after each convolutional layer (eg. The input size is $512\times 512$, after one convolutional layer the size becomes $256\times 256$) while the number of filters (i.e. number of feature maps) are doubled from 32 all the way to 512.

Up to now, our approach considers to learn a generic representation from a (usually limited) set of training examples, which is then employed in generating filamentary structured phantoms, which is also referred to as \emph{Fila-GAN}. Next, we consider a variant inspired by the recent style transfer techniques.

\subsection{Fila-sGAN: the Style transfer variant}
Inspired by the recent advance in image style transfer, here we consider an alternative variant of our approach as illustrated in Fig.~\ref{fig:flowchart}(a) (now including the components in the dashed box): Given an input segmentation $\vec{y}$ that delineates its filamentary structured content, the generated phantom $\vec{\hat{x}}$ is expected to possess the distinct style (i.e. texture) of a target style input $\vec{x}_{\mathrm{s}}$, while still adhering to the content of $\vec{y}$ that has been presented during training stage. This variant is also termed \emph{Fila-sGAN}, to highlight the fact that the synthesized image is now based on a \emph{particular} style representation provided by $\vec{x}_{\mathrm{s}}$, instead of the generic representation as we have aimed for in our basic approach \emph{Fila-GAN}. This is made possible by introducing a style image as an additional training input, i.e. a new filamentary structured image $\vec{x}_{\mathrm{s}}$ that comes with a different style or texture. Note in general $\vec{x}_{\mathrm{s}}$ has its own filamentary structure (aka segmentation), which is usually different from the other input argument $\vec{y}$, nonetheless this should not affect the synthesis performance of our approach.
It is worth noting that this design make practical sense in biomedical imaging setting: On one hand very few images are annotated, while on the other hand, there could be abundant un-annotated images out there that could serve as potential style inputs.

The training and testing processes of this variant largely remain the same as what we have described in our approach: The training is still carried out in batch mode over the $n$ annotated training examples. The aforementioned generator and the discriminator functions still carry on. The differences lie on the objective function of Eq.~\eqref{eq:minmax}, where the last term $\lambda L_{\mathrm{DEV}}\left(G_{\theta}\right)$ is now replaced by a new cost term $L_{\mathrm{ST}}\left(G_{\theta}\right)$ to be defined below.
%
Without loss of generality, we follow the style transfer idea of~\cite{UlyEtAl:icml16,JusEtAl:ECCV16} to utilize the convolutional neural network (CNN) of VGG-19~\cite{simonyan2014very} to extract features from its multiple layers.
Accordingly to the network architecture of VGG-19, it can be represented as a series of five CNN blocks of the VGG net, with each block containing 2-4 consecutive convolution layers of the same size.
For notation convenience, let $\Gamma$ indexes a set of CNN blocks, and for a particular block $\gamma \in \Gamma$, its set of layers is represented by $\Lambda(\gamma)$ or simply $\Lambda$, with a layer index being $\lambda \in \Lambda$. Now, for a filamentary structured image $\vec{X}$ (being either the real $\vec{x}$ or phantom $\vec{\hat{x}}$), denote by $\phi_{\gamma}^{(\lambda)}\left(\vec{X}\right)$ the $\lambda$-th corresponding layer of CNN block $\gamma$.
The VGG-19 net is obtained by training on the ImageNet image classification task, with its further details being displayed in Fig.~\ref{fig:flowchart}(d).
In terms of the resulting optimization problem, it is realized in our approach by explicitly incorporating two perceptual loss components of~\cite{GatEckBet:arxiv15}, namely the style loss and the content loss, as well as a total variation loss, which are to be described next.

\paragraph*{\textbf{Style loss}}
The style loss is used to minimize the textural deviation between the target style $\vec{x}_{\mathrm{s}}$ and the phantom $\vec{\hat{x}}$. To achieve this, consider $\Gamma_{\mathrm{s}}$ indexing the set of CNN blocks, and for each block index $\gamma_{\mathrm{s}} \in \Gamma_{\mathrm{s}}$, its set of layers being represented by $\Lambda_{\mathrm{s}}$.
Now, define the $\lambda_{\mathrm{s}}$-th layer of block $\gamma_{\mathrm{s}}$, $\phi_{\gamma_{\mathrm{s}}}^{(\lambda_{\mathrm{s}})} \left(\vec{X}\right)$, where $\vec{X}=\vec{x}_{\mathrm{s}}$ or $\vec{X}=\vec{\hat{x}}$. With a slight abuse of notation, denote by $\left|\lambda_{\mathrm{s}}\right|$ the number of feature maps in current layer $\lambda_{\mathrm{s}}$. Let $i$ (or $j$) index the feature map of interest, and let $k$ index an element of the current feature map $i$ (or $j$). The corresponding feature information is characterized by the \emph{Gram matrix} $\mathbb{G}_{\gamma_{\mathrm{s}}}^{(\lambda_{\mathrm{s}})} \left(\vec{X}\right) \in \mathbb{R}^{\left|\lambda_{\mathrm{s}}\right| \times \left|\lambda_{\mathrm{s}}\right|}$, where each element $\mathbb{G}_{\gamma_{\mathrm{s}}, ij}^{(\lambda_{\mathrm{s}})}$ defines an inner product between the $i$-th and $j$-th feature maps in $\lambda_s$-th layer of $\gamma_s$-th block:
\begin{align}
\label{eq:gram_max}
\mathbb{G}_{\gamma_{\mathrm{s}}, ij}^{(\lambda_{\mathrm{s}})} = \sum_{k} \phi_{\gamma_{\mathrm{s}}, ik}^{(\lambda_{\mathrm{s}})} \phi_{\gamma_{\mathrm{s}}, jk}^{(\lambda_{\mathrm{s}})}.
\end{align}
Now the style loss between $\vec{x}_{\mathrm{s}}$ and $\vec{\hat{x}}$ during training becomes
{\small
\begin{align}
\label{eq:style_loss}
l_{\mathrm{sty}} \left( G_{\theta} \right) = \sum_{\gamma_{\mathrm{s}} \in \Gamma_{\mathrm{s}}, \lambda_{\mathrm{s}} \in \Lambda_{\mathrm{s}}} \frac{\varpi_{\gamma_{\mathrm{s}}}}{W_{\gamma_{\mathrm{s}}} H_{\gamma_{\mathrm{s}}}} \left\| \mathbb{G}_{\gamma_{\mathrm{s}}}^{(\lambda_{\mathrm{s}})}\left(\vec{x}_{\mathrm{s}}\right) - \mathbb{G}_{\gamma_{\mathrm{s}}}^{(\lambda_{\mathrm{s}})}\left(\vec{\hat{x}}\right) \right\|_{\mathrm{F}}^2,
\end{align}
}
where $\|\cdot\|_{\mathrm{F}}$ is the matrix Frobenius norm, $\varpi_{\gamma_{\mathrm{s}}}$ denoting the weight of $\gamma_{\mathrm{s}}$-th block \textit{Gram matrix}. Note $\vec{\hat{x}} = G_{\theta} \left( \vec{y}, \vec{z} \right)$ by definition.

\paragraph*{\textbf{Content loss}}

For content loss, consider the following notation: Let $\Gamma_{\mathrm{c}}$ index the set of CNN blocks, and for each block index $\gamma_{\mathrm{c}} \in \Gamma_{\mathrm{c}}$, its set of layers is denoted by $\Lambda_{\mathrm{c}}$.
As already stated, the synthesized phantom $\vec{\hat{x}}$ is expected to abide by the filamentary structure as prescribed in the real raw image $\vec{x}$ of the segmentation input, which is carried out by encouraging them to match up, i.e. by minimizing the following Frobenius norm of the difference between input and generated CNN features:
\begin{align}
\label{eq:content_loss}
l_{\mathrm{cont}} \left( G_{\theta} \right) = \sum_{\gamma_{\mathrm{c}} \in \Gamma_{\mathrm{c}}, \lambda_{\mathrm{c}} \in \Lambda_{\mathrm{c}}} \frac{1} {W_{\gamma_{\mathrm{c}}} H_{\gamma_{\mathrm{c}}}} \left\| \phi_{\gamma_{\mathrm{c}}}^{(\lambda_{\mathrm{c}})} \left(\vec{x}\right) - \phi_{\gamma_{\mathrm{c}}}^{(\lambda_{\mathrm{c}})} \left(\vec{\hat{x}}\right) \right\|_{\mathrm{F}}^2.
\end{align}

\paragraph*{\textbf{Total variation loss}}
Furthermore, we consider encouraging spatial smoothness in the generated phantom by incorporating the following total variation loss:
\begin{align}
\label{eq:tv_loss}
l_{tv} \left( G_{\theta} \right) = \sum_{w,h} \left( \left\|\hat{x}_{w,h+1} - \hat{x}_{w,h}\right\|_2^2 + \left\|\hat{x}_{w+1,h} - \hat{x}_{w,h}\right\|_2^2 \right),
\end{align}
with $w,h \in W,H$, and $\hat{x}_{w,h}$ denotes the pixel value of given location in phantom image $\vec{\hat{x}}$.

The above-mentioned three loss functions together leads to
$L_{\mathrm{ST}} \left( G_{\theta} \right) = w_{\mathrm{cont}} l_{\mathrm{cont}} + w_{\mathrm{sty}} l_{\mathrm{sty}} + w_{\mathrm{tv}} l_{\mathrm{tv}}$.
%
Thus we consider the above style transfer loss $L_{\mathrm{ST}}$, instead of $L_{\mathrm{DEV}}$ as in Eq.~\eqref{eq:minmax}.
Accordingly, generator $G$ now takes the objective function of the following form
\begin{align}
\label{eq:g_loss_style}
L_{\mathcal{G}}(G_{\theta}) = - \sum_i \log D_{\gamma} \left(G_{\theta}\left(\vec{y}_i,\vec{z}\right)\right) + L_{\mathrm{ST}}(G_{\theta}),
\end{align}
while the objective function of discriminator $D$ of Eq.~\eqref{eq:d_loss} remains unchanged. It is clear that in this variant, the style transfer contribution from the target style $\vec{x}_{\mathrm{s}}$ is obtained by back-propagation optimization of the above objective function, while the rest ingredients of our approach are kept as the same.

\section{Experimental details}

\paragraph*{\textbf{Datasets and Preparation}}
Empirically our approach is examined on four standard benchmarks that covers a broad spectrum of filamentary structured images including both retinal blood vessels and neurons. They are DRIVE~\cite{StaEtAl:tmi04}, STARE~\cite{HooEtAl:tmi00}, high-res fundus or HRF~\cite{kohler2013automatic}, as well as 2D Neurons or NeuB1~\cite{DeEtAl:TMI15}. The image sizes and the amount of training examples are also rather different across these datasets:
DRIVE contains 20 training examples and 20 test images, with each of size $584\times 565$.
STARE and HRF are two fundus image datasets with image sizes being $700\times 605$ and $3,304\times 2,336$ respectively, as well as split of training/testing images being 10/10 and 22/23 respectively.
The NeuB1 dataset is considered here for microscopic neuronal images, which contains 112 images of size $512\times 512$. We also follow the standard train/test split of 37/75 as in~\cite{DeEtAl:TMI15}.

To summarize, the image sizes of DRIVE, STARE, and NeuB1 are roughly similar, while HRF contains high resolution (and subsequently much larger size) images.
Then in preprocessing, raw images of these first three datasets are all resized to a standard size of $512\times 512$, as follow: For DRIVE, as all images are of size $584 \times 565$ and contain a relatively large-size background area (determined if a pixel is outside of a prescribed circular-shaped mask), they are cropped into $565 \times 565$ sub-images centered around the original ones, that ensures all foreground pixels are still contained in the cropped images. They are further resized to $512 \times 512$ by bicubic interpolation;
For STARE, as the images possess rather small background margins outside of its foreground mask, they are directly resized to the same size of $512\times 512$ by bicubic interpolation;
Now for HRF, as its raw images are of much larger size of $3304\times 2336$, these images are all resized to $2,048\times 2,048$ instead of the $512\times 512$ template size we have used previously, in order to retain sufficient information of the original images.
The pixel values of all input image signals are scaled to the same range of $[-1, 1]$, for each of the benchmarks. 
As a consequence, the learned generator in our approach will produce a phantom image of size $512\times 512$ for DRIVE, STARE, and NeuB1, and size $2,048\times 2,048$ for HRF, respectively. These images are subsequently upsampled to their original sizes. For any of the above-mentioned fundus image datasets, the final results are obtained by applying its prescribed circular-shaped mask, so only inside pixels are retained as foreground.

\paragraph*{\textbf{Model Architecture and internal parameters}}
Fig.~\ref{fig:flowchart}(b-d) displays the architecture of our approach: In the generator and discriminator modules, each 3D box denotes a CNN layer consisting its feature maps,
a directed edge usually represents a convolutional (or deconvolutional) operation with a filter size $w_{\mathrm{f}} \times h_{\mathrm{f}} \times l_{\mathrm{f}}$.
In this paper we consider $w_{\mathrm{f}}=h_{\mathrm{f}}=4$ pixels with $l_{\mathrm{f}}$ self-manifested by the third dimension of its consecutive layer.
Note Fig.~\ref{fig:flowchart}(b-c) specify the intrinsic parameter values of our networks $G$ and $D$, such as the size of each layers, and the length of the noise code $Z=400$.
In particular, in generator $G$, the $\bigoplus$ sign together with the two directed edges pointing to it denote a concatenation operation.
For example,
the first $\bigoplus$ sign shows a concatenation operated on a $8 \times 8 \times 512$ tensor and a $8 \times 8 \times 256$ tensor that produces a $8 \times 8 \times 768$ tensor;
This is followed by a deconvolutional operation of filter size $4 \times 4 \times 512$ that produces the 3D box of size $16 \times 16 \times 512$.

Throughout the experiments in our approach, the following internal parameters are empirically considered:
The TensorFlow deep learning library is used with training epochs being fixed to 100.
To initialize the neural net weights of discriminator $D$ and generator $G$ (i.e. parameters $\theta$, $\gamma$), a [-0.04, 0.04] truncated normal distribution of zero-mean and standard deviation of 0.02 is engaged. The weights $\theta$ of $G$ is updated with mini-batch of size 1 using Adam optimizer~\cite{kingma2014adam}. Meanwhile, the vanilla stochastic gradient descent is employed for $D$ to update $\gamma$. Learning rate is set to 0.0002 for the generator and 0.0001 for the discriminator during back-propagation training. $\lambda$ in our \emph{Fila-GAN} is set to 100.
To balance the learning progress of $G$ and $D$, we choose to update $G$ twice then update $D$ once during each learning iteration.
During training, the noise code is sampled elementwise from zero-mean Gaussian with standard deviation 0.001;
At testing run, it is sampled in the same manner but with a different standard deviation of 1. Empirically this is found useful in maintaining proper level of diversity for our small sample-size situation.
We follow the practice of~\cite{RadMetChi:arxiv15} and choose not to apply batch normalization to the output layer of $G$ as well as to the input layer of $D$.
Otherwise, batch normalization~\cite{ioffe2015batch} is utilized right after each convolutional layer, as overall better model training behaviors have been observed for both $G$ and $D$ nets.

For our style transfer variant, \emph{Fila-sGAN}, the VGG-19 nets are employed to produce the feature descriptors.
%
Some of their layers are used for extracting style/content features, which are illustrated in Fig.~\ref{fig:flowchart}(d), and are specified as follows:
The sets of block indices for style and content losses are $\Gamma_s =\{1,2,3,4,5\}$, and $\Gamma_c =\{4\}$, respectively.
Besides, for any block $\gamma_s \in \Gamma_s$, its particular set of layer indices is $\Lambda_s (\gamma_s)=\{1\}$ for the style loss, and $\Lambda_c=\{2\}$ for the content loss. $\varpi_{\gamma_{\mathrm{s}}}$ is fixed to $0.2$ over all blocks in $\Gamma_s$.
Meantime, the weights of the three respective loss functions, namely $w_{\mathrm{cont}}, w_{\mathrm{sty}}, w_{\mathrm{tv}}$, are 1, 10, and 100, accordingly.

\paragraph*{\textbf{Computation Time}}
All the experiments are carried out on a standard PC with Intel iCore 7 CPU and Titan-X GPU with 12GB memory.
Our implementation of the proposed \emph{Fila-GAN} and \emph{Fila-sGAN} is in Python.
Training time of \emph{Fila-GAN} and \emph{Fila-sGAN} on DRIVE takes around 108 minutes and 184 minutes, respectively.
The average run-time speed on synthesizing a DRIVE-size image is 0.4471s. 

\begin{figure}
\includegraphics[width=0.99\linewidth]{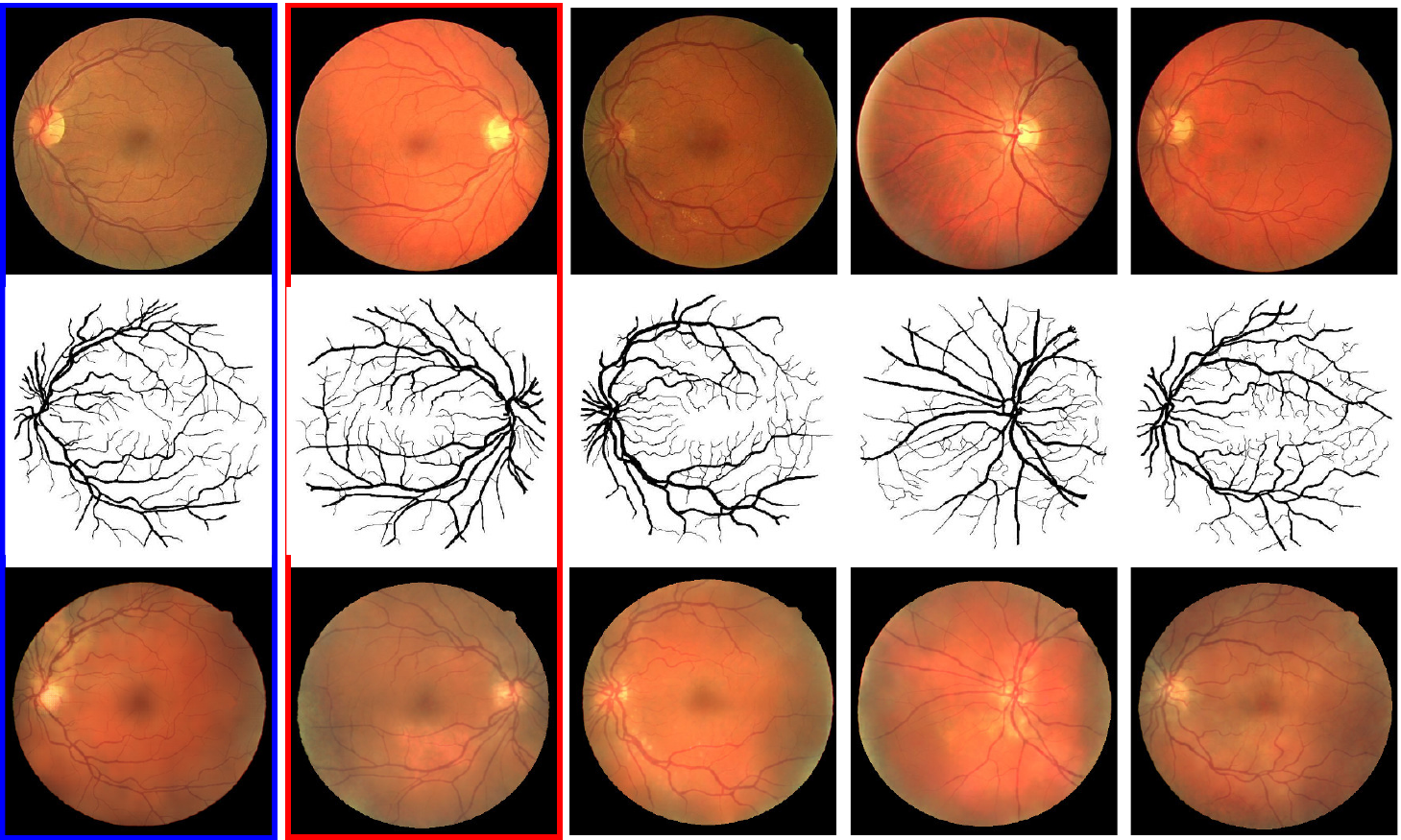}
\centering
\caption{\footnotesize{Exemplar DRIVE phantoms generated by \emph{Fila-GAN}. For each column, 1st row presents a real image, 2nd is the corresponding ground-truth, 3rd displays one generated phantom.}}
\vspace{-3mm}
\label{fig:visual_result_drive}
\end{figure}

\begin{figure}
\centering	
\includegraphics[width=0.99\linewidth]{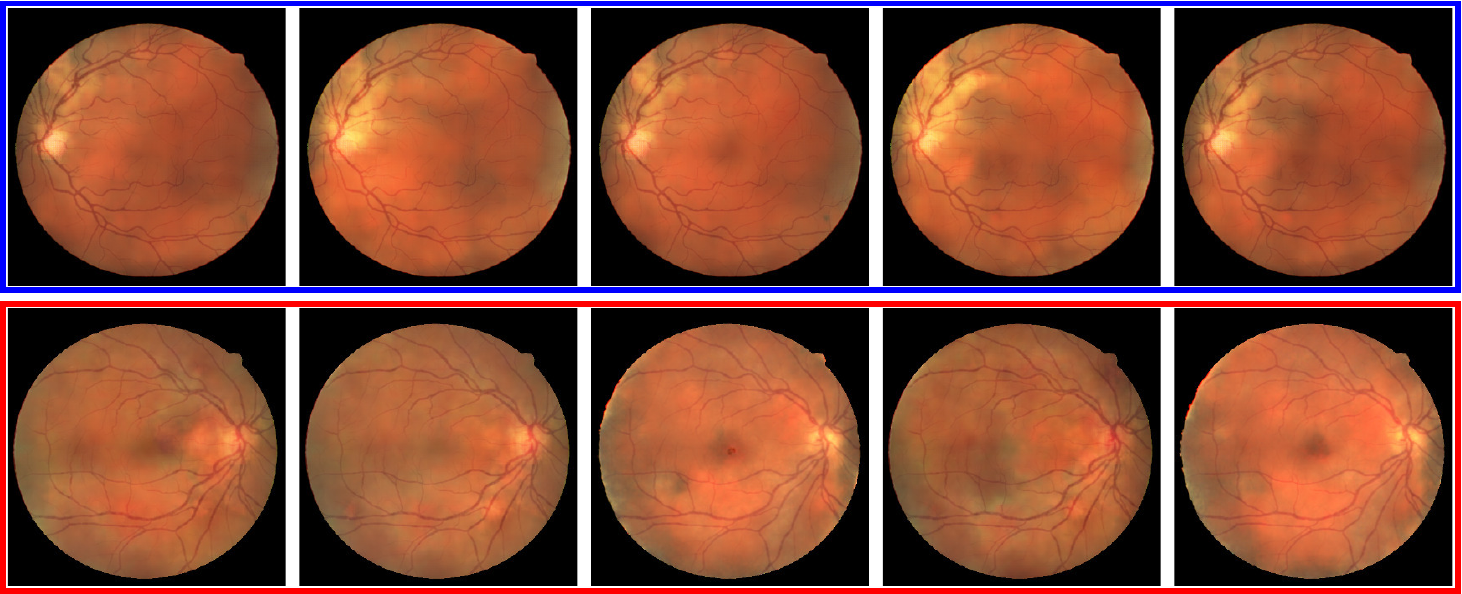}
\caption{\footnotesize{The two rows in this figure display multiple synthesized results of the first two columns of Fig.~\ref{fig:visual_result_drive} respectively, to showcase the ability of our approach in generating diverse outputs.}}
\vspace{-5mm}
\label{fig:div_result}
\end{figure}

\begin{figure}
\vspace{-3mm}
\centering
\includegraphics[width=0.99\linewidth]{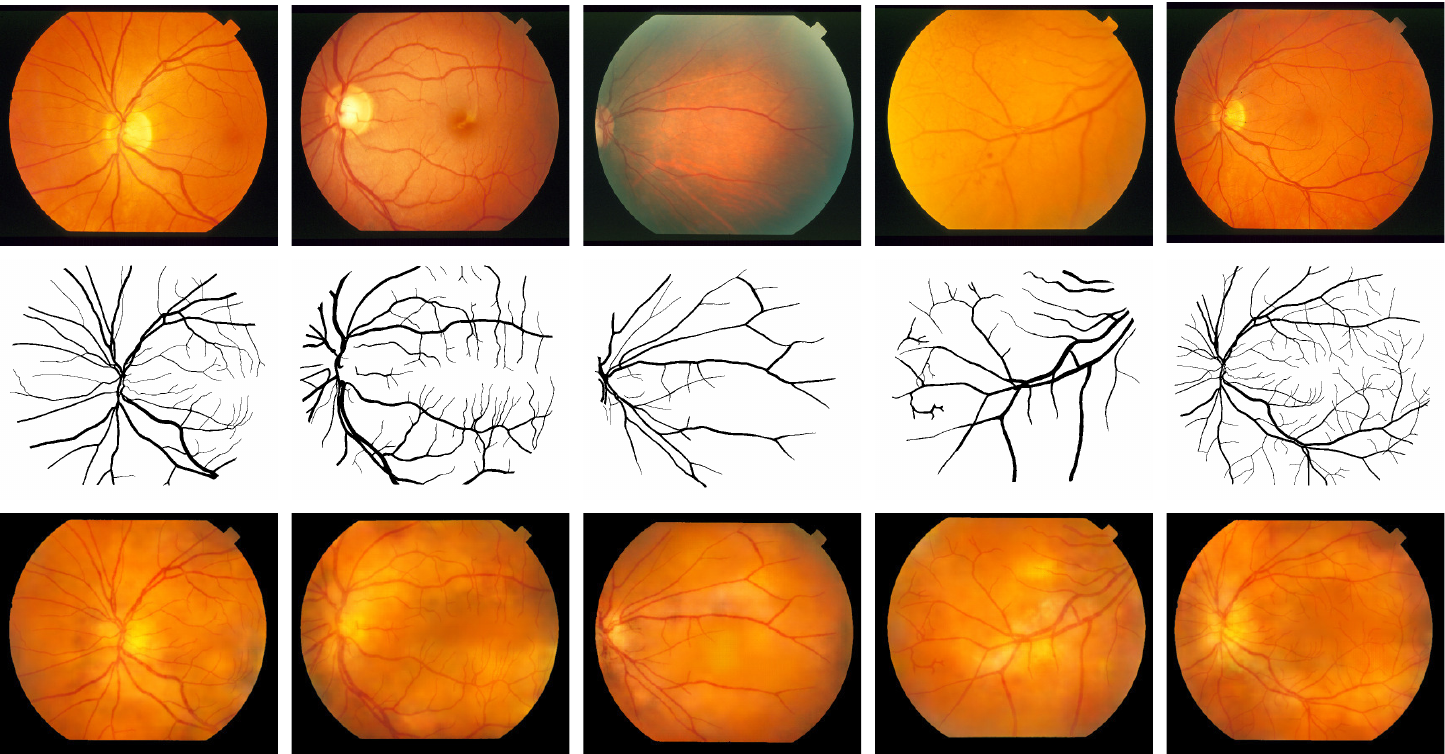}
\caption{\footnotesize{Exemplar STARE phantoms generated by \emph{Fila-GAN}. For each column, 1st row presents a real retinal image, 2nd is the corresponding ground-truth, while 3rd displays one generated phantom.}}
\vspace{-5mm}
\label{fig:visual_result_stare}
\end{figure}

\begin{figure}
\centering
\includegraphics[width=0.99\linewidth]{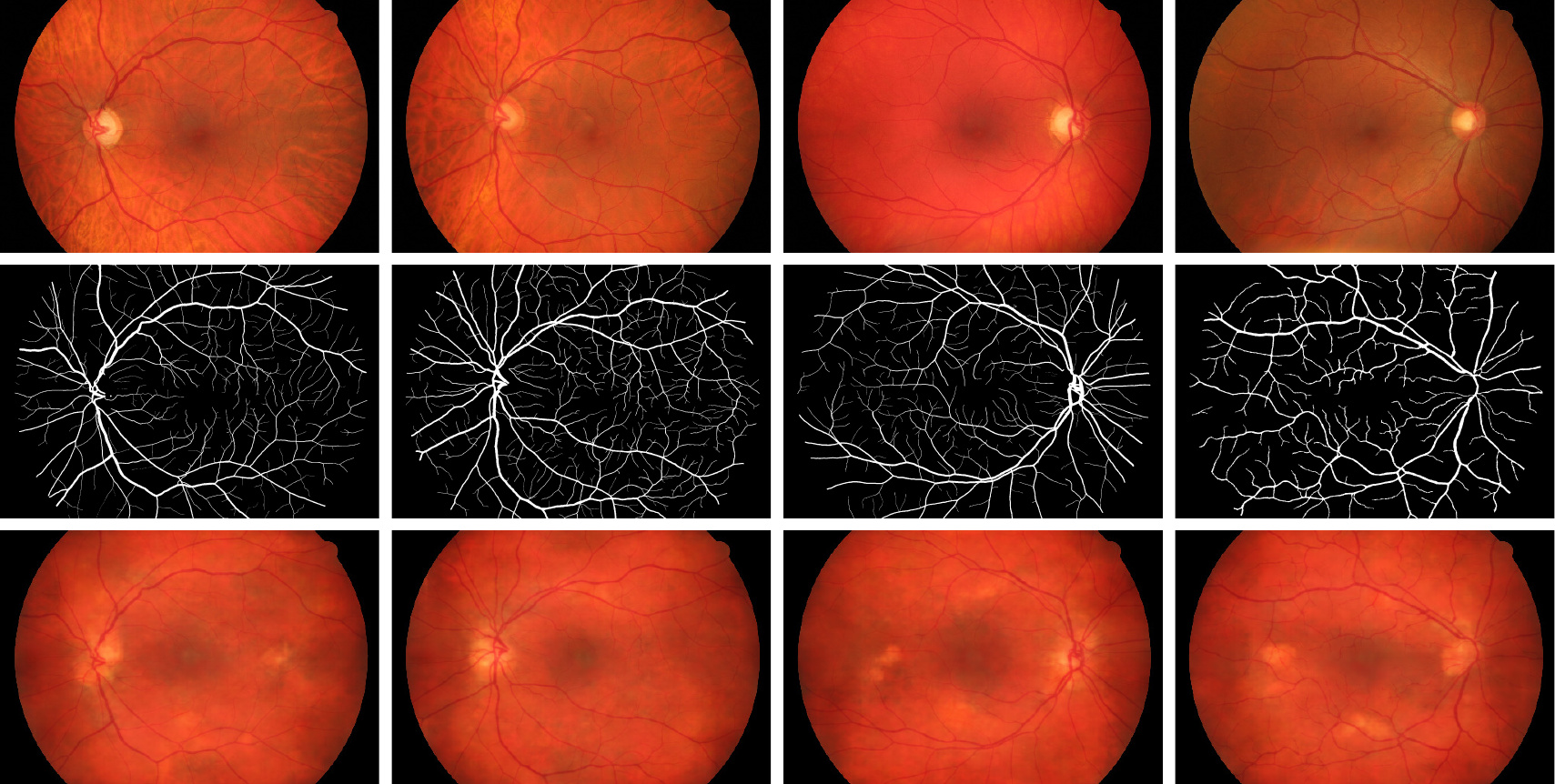}
\caption{\footnotesize{Exemplar HRF phantoms generated by our \emph{Fila-GAN} model. For each column, 1st row presents a real retinal image, 2nd is the corresponding ground-truth, while 3rd displays one generated phantom. A zoomed-in figure is presented in the supplementary file.}}
\vspace{-5mm}
\label{fig:visual_result_hrf}
\end{figure}

\begin{figure}
\centering
\includegraphics[width=0.99\linewidth]{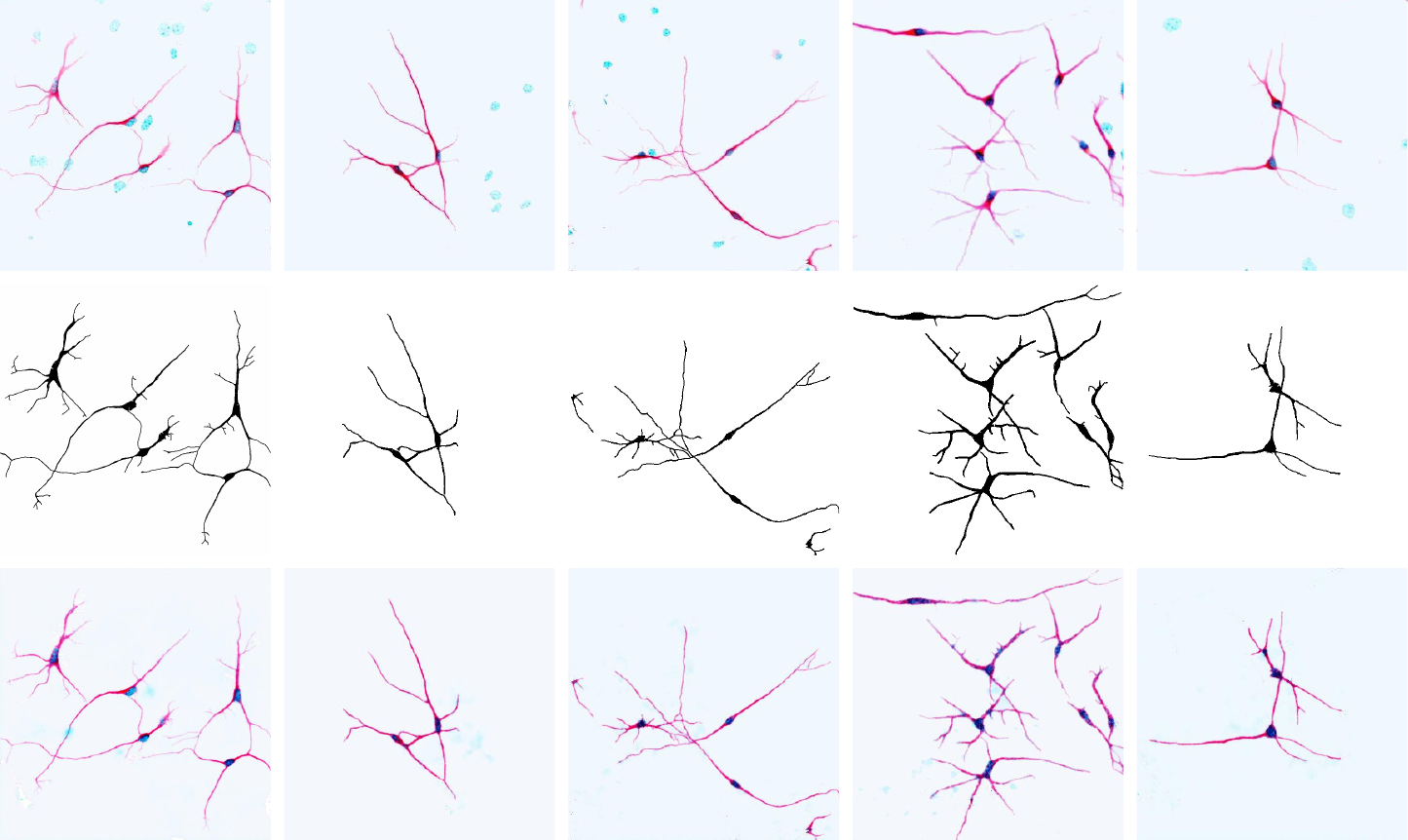}
\caption{\footnotesize{Exemplar NeuB1 phantoms generated by our \emph{Fila-GAN}. For each column, 1st row presents a real neuronal image, 2nd the corresponding ground-truth, while 3rd displays one generated phantom.}}
\vspace{-5mm}
\label{fig:visual_result_nb1}
\end{figure}

\begin{figure}
\centering
\includegraphics[width=0.99\linewidth]{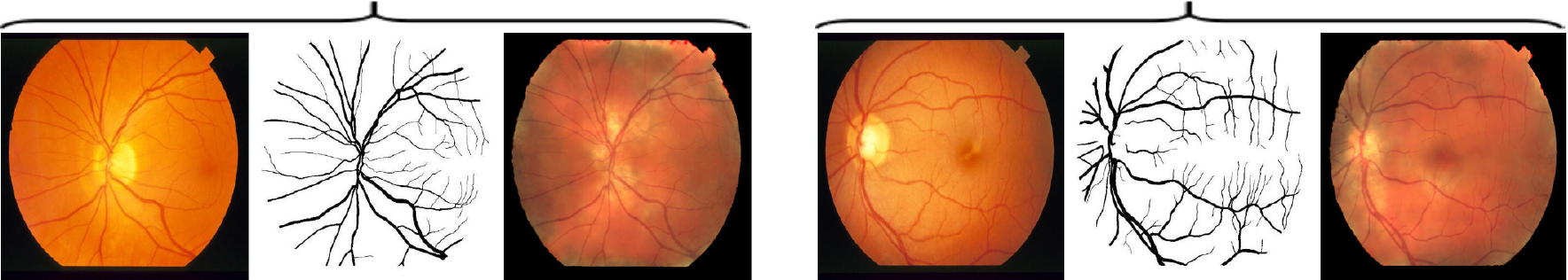}
\caption{\footnotesize{Apply the \emph{Fila-GAN} model trained on DRIVE training set (the same model used in Fig.~\ref{fig:visual_result_drive}) to STARE segmentation maps. Two random samples are selected, from left to right are real STARE image, ground-truth, and generated phantom, respectively.}}
\vspace{-5mm}
\label{fig:visual_result_stare_from_drive}
\end{figure}

\begin{figure}
\centering
\includegraphics[width=0.99\linewidth]{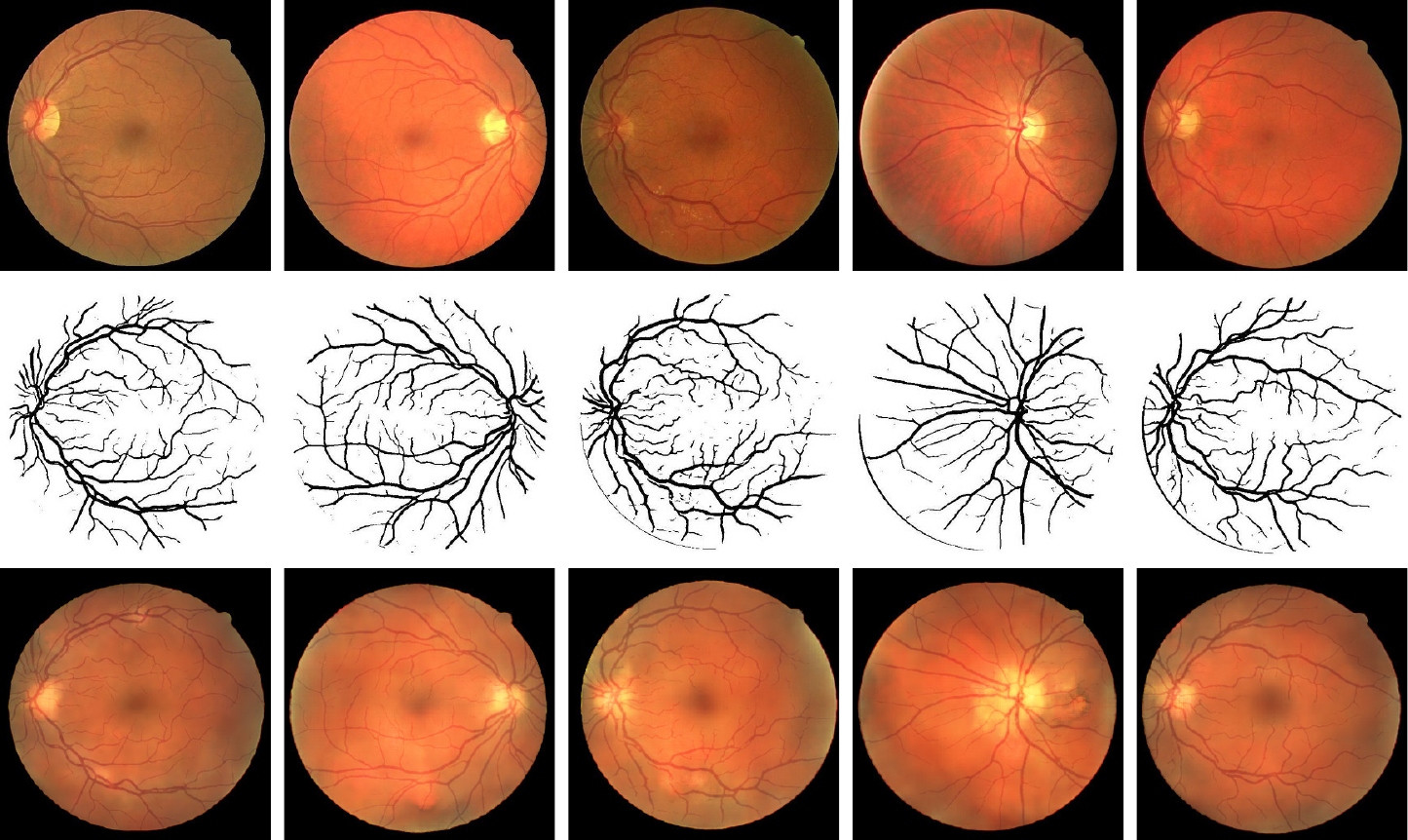}
\caption{\footnotesize{Exemplar phantoms generated from segmentation predictions (instead of ground-truths) of a typical segmentation method trained on DRIVE. In each column, 1st row presents a real DRIVE image, 2nd the corresponding ground-truth, 3rd displays one generated phantom.}}
\vspace{-4mm}
\label{fig:visual_result_drive_seg}
\end{figure}

\newpage

\begin{figure*}
\centering
\includegraphics[width=0.9\linewidth]{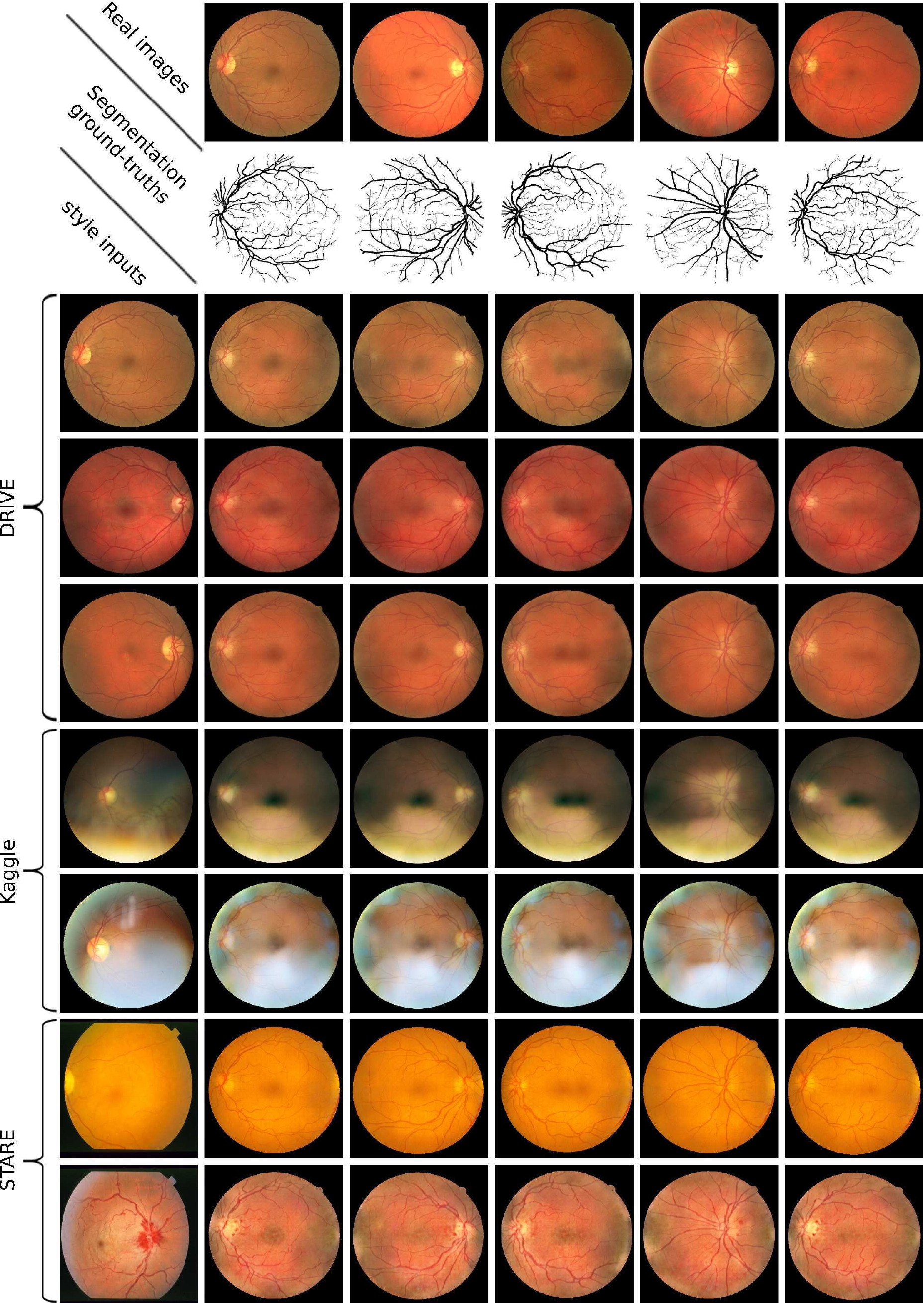}
\caption{\footnotesize{Phantoms generated by \emph{Fila-sGAN}. The first and second rows display the real DRIVE images in the hindsight, and corresponding ground-truths, respectively. From third row onwards, each row presents the phantoms synthesized from a specific style image shown in the first column. See text for details.}}
\label{fig:visual_result_style}
\end{figure*}

\section{Empirical Experiments}

\subsection{Visual results}
Fig.~\ref{fig:visual_result_drive} displays exemplar synthesized results of applying \emph{Fila-GAN} on DRIVE, where the first row presents the real images, the second row refers to the corresponding ground-truth, and the last row displays the generated phantoms. It can be observed that the phantoms preserve the vascular morphology of the input (the second row in figure), while being able to present different yet realistic-looking texture appearances. It is interesting to note that the very bright colored areas are usually properly situated around the optical disk locations of the ground-truth images, which suggests that our phantom generation model could capture such intrinsic correlations without explicit human interventions for conveying such prior knowledge. Similar results can also be found in other datasets, e.g. STARE (Fig.~\ref{fig:visual_result_stare}), HRF (Fig.~\ref{fig:visual_result_hrf}), and neuronal dataset NeuB1 (Fig.~\ref{fig:visual_result_nb1}).
%
To demonstrate the strength of \emph{Fila-GAN} in generating multiple distinct syntheses from the same ground-truth, Fig.~\ref{fig:div_result} further presents more synthesized results of the first and second ground-truth inputs as of Fig.~\ref{fig:visual_result_drive}, obtained by randomly drawing i.i.d. noise codes $\vec{z}$. Clearly, for each of these ground-truth input, the results synthesized by \emph{Fila-GAN} are visually different. It can be observed that \emph{Fila-GAN} is relatively more powerful in emulating textural diversity and less in capturing illumination changes.
In addition, this \emph{Fila-GAN} model trained on DRIVE training set is applied to images from other datasets (here we consider the STARE dataset~\cite{HooEtAl:tmi00}), with exemplar results presented in Fig.~\ref{fig:visual_result_stare_from_drive}: Two random samples are selected, from left to right are real image, ground-truth, and phantom, respectively. Not surprisingly, the two synthesized STARE phantoms bear DRIVE-like textures.
%
Finally, to demonstrate that \emph{Fila-GAN} works well even without ground-truth segmentation, an experiment is carried out to evaluate its behavior when a typical segmentation method is employed to produce binary segmentation maps.
Note in fact any reasonable segmentation method should work, while we consider here a baseline convolutional neural nets segmentation method whose details are described in the supplementary file.
As displayed in Fig.~\ref{fig:visual_result_drive_seg}, the generated phantoms are visually also plausible. 
We note in the passing that promising results are also obtained on the BigNeuron 3D neuronal image stacks~\cite{PenEtAl:neuron15} as discussed in the supplementary file.

Now we switch gear to examine our style transfer variant, \emph{Fila-sGAN}.
Fig.~\ref{fig:visual_result_style} presents a gallery of collective results to the same set of DRIVE retinal images (shown in columns) when trained with different style images (shown in rows).
Here the real images of hindsight are presented in the first row, followed by their corresponding ground-truths in the second row. From 3rd row onwards, the generated phantoms with different styles are presented.
DRIVE, Kaggle\footnote{Kaggle images are obtained from~\url{https://www.kaggle.com/c/diabetic-retinopathy-detection}, a contest for diabetic retinopathy detection.}, STARE denotes the three distinct style sources.
It is observed that the texture style of each phantom synthesized by our \emph{Fila-sGAN} is clearly controlled by its particular style input, which is especially pronounced for the Kaggle style images that are considerably different from the rest images. Meanwhile the respective filamentary structures are well preserved.
Moreover, visually diverse results are again generated by varying the noise input $\vec{z}$, as is presented in 
Fig.4 of the supplementary file.
Visually \emph{Fila-sGAN} is shown to be capable of producing realistic looking phantoms of very different styles.

\subsection{Quantitative results}
It is often difficult to quantitatively evaluate the quality of synthesized results, which is also mentioned in~\cite{IsoEtAl:arxiv16,salimans2016improved}.
We start by assuming to have access to any reasonable supervised segmentation method (here we consider the aforementioned baseline CNN segmentation method).
In our context, we consider two relatively straightforward evaluation schemes:
Scheme 1 is to examine the usefulness of these newly generated images in boosting the performance of the same segmentation method;
Scheme 2 is through investigating the differences in segmentation results, when applying the same trained segmentation model on both synthesized and real retinal images.

%
%
%
%

\begin{table}
\centering
\caption{\footnotesize{Quantitative results of evaluation scheme 1. Comparison of segmentation performance of a baseline segmentation method on the same DRIVE, STARE, HRF, NeuB1 test set while training on a different set of images (as depicted in the left-most column). Results are evaluated by average F1-score (\%)}. See text for details.}
\label{tab:syn_seg_result}
\begin{tabular}{|c|c|c|c|c|}
\hline
 & DRIVE & STARE & HRF &  NeuB1\\
\hline
\emph{synthetic images}  & 71.44 	 		& 75.34 		& 69.28					 & 77.69 \\
\hline
\emph{real images}  & 79.15 	 		& 78.11		& 78.68				 & 83.91 \\
\hline
\emph{real + synthetic images}  & 80.33 	 		& 79.02		& 79.50					 & 85.06 \\
\hline
2\emph{$\times$ real images}  & 80.70	 		& 79.66		& 79.77					 & 85.40 \\
\hline
\end{tabular}
\vspace{-4mm}
\end{table}

\paragraph*{Evaluation Scheme 1}
As our baseline CNN segmentation method takes image patches as input, we start by preparing ready 800K image patches from real images in the training set which are evenly partitioned into the first 400K and the second 400K real image patches. Additionally we have another 400K image patches from the synthesized training images. This gives rise to 4 scenarios, which are: \emph{synthetic images} where the segmentation baseline is trained on the 400K synthetic patches; \emph{real images} where it is trained on the first set of 400K real image patches; \emph{real + synthetic images} for training on the first set of 400K real patches plus the 400K synthetic patches; \emph{2$\times$ real images} where the training is on both sets of 400K real patches. As summarized in Table~\ref{tab:syn_seg_result}, the segmentation model performs the worst when training only on synthetic images, which is to be expected.
However, for example on DRIVE, compared with $79.15\%$ F1-score when training on 400K real patches, the introduction of additional synthetic images, i.e. \emph{real + synthetic images}, is demonstrated to improve the segmentation performance to $80.33\%$. As a side note, we observe that when replacing synthetic images by real ones, i.e. \emph{2$\times$ real images}, there is a further boost in segmentation performance to $80.70\%$. It suggests that overall synthesized images generated by \emph{Fila-GAN} helps to improve segmentation performance; Meanwhile it is still preferable if more real data are available, which is to be expected.
The same trend is consistently presented when working with all these different benchmarks.
Further, similar empirical observations have also been made for our \emph{Fila-sGAN} variant. Take the DRIVE benchmark for example, as indicated in Table~1 
of the supplementary file, similar patterns as presented in Table~\ref{tab:syn_seg_result} for \emph{Fila-GAN} also hold true for \emph{Fila-sGAN}, where the segmentation results of \emph{real + synthetic images} leads to a boost of performance to $80.49\%$.
%
Finally, to put our segmentation results into context, we also cite the comparable average F1 score results of the state-of-the-art segmentation methods including
SF-context distance~\cite{GuEtAl:TMI16} (78.86\%), DRIU~\cite{ManEtAl:MICCAI16} (82.20\%), among others on the DRIVE benchmark.


\begin{figure}
\centering	
\includegraphics[width=0.99\linewidth]{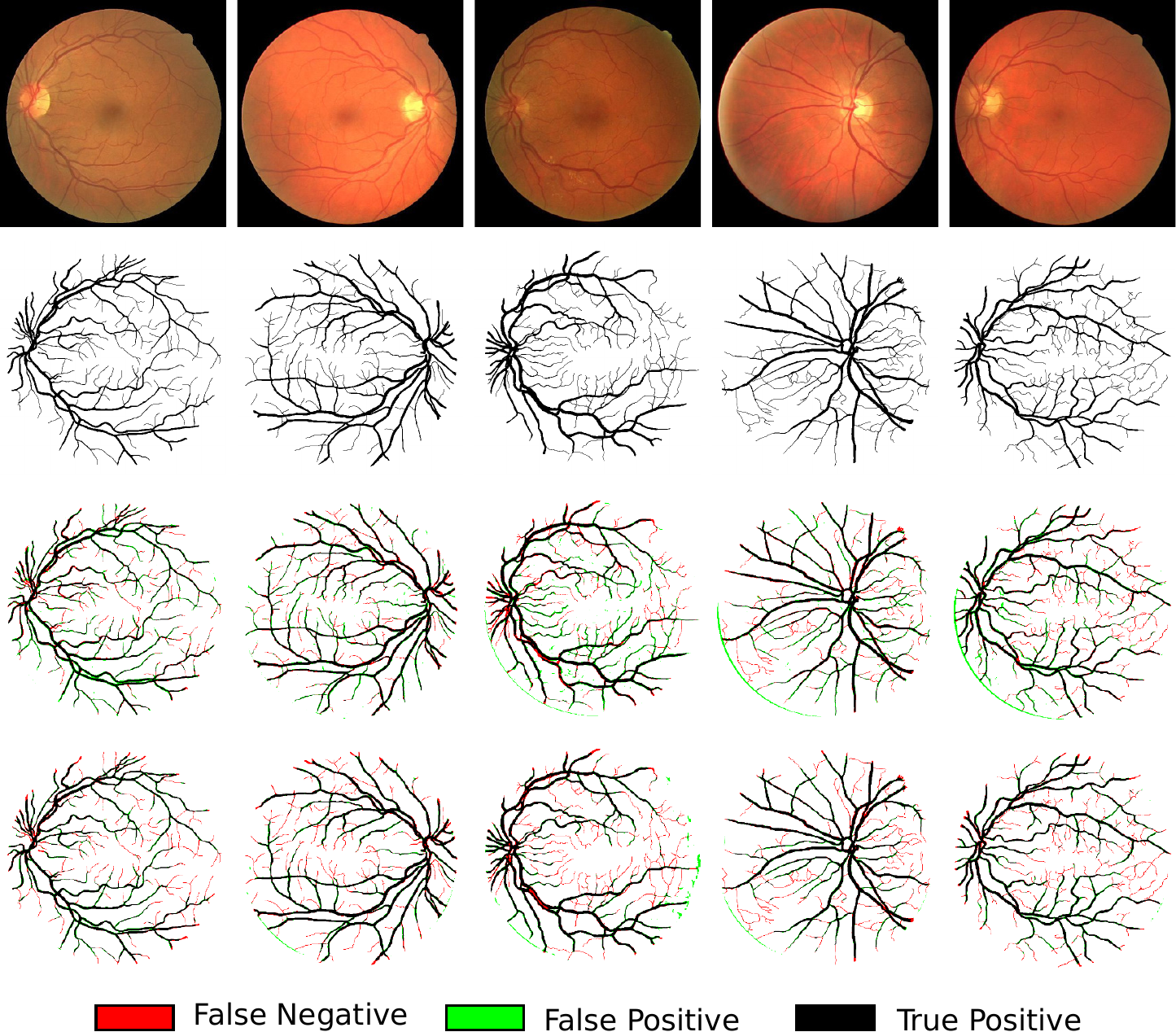}
\caption{\footnotesize{visual comparison on segmentation results of real vs. synthetic images on DRIVE test set images (i.e. evaluation scheme 2). First row displays the real images, with the corresponding ground-truths shown in the second row. Third and fourth rows then present segmentation results of the real and synthesized phantom images, respectively. Note these phantom images are exactly the same images shown in Fig.~\ref{fig:visual_result_drive}.
Here \textcolor{black}{black} refers to true positive (TP), \textcolor{green}{green} stands for false positive (FP), and \textcolor{red}{red} is false negative (FN).}}
\vspace{-5.5mm}
\label{fig:syn_seg_compare}		
\end{figure}


\paragraph*{Evaluation Scheme 2}
We follow the second evaluation scheme and inspect the segmentation results of real vs. generated phantoms when the same segmentation model trained on the real training set is employed.
The segmentation method is firstly trained on the training set of the current benchmark in consideration, which is then applied on both real test set images and phantoms generated from the corresponding ground-truths. The widely used average F1-score is considered as our segmentation evaluation metric.
%
Table~2 of the supplementary file summarizes quantitative results across the benchmark datasets of retinal and neuronal images. That quantitative performance on these two test sets (real vs. phantom) are indeed similar w.r.t. the same segmentation model. We attribute the relatively higher performance on the phantom images to their perhaps more homogeneous textural appearances.
%
Visual comparison is also provided in Fig.~\ref{fig:syn_seg_compare}.
Given that the segmentation model is trained on the real training set, by and large the vascular structures are still well-segmented when applying to the phantom images. Compare with the segmentation results of real images, the phantoms tend to incur relatively less false positive pixels at the price of relatively more missing vessel pixels that are usually those tiny and thin filaments.
It is also observed that segmentation results of different synthetic images based on the same segmentation input seem to vary little, and the differences are mostly concentrated on these thin and tiny filaments.

\section{Conclusion}
We propose a novel data-driven approach to synthesize filamentary structured images given a ground-truth input. The synthesized images are realistic-looking, and have been shown to boost image segmentation performance when used as additional training images. Moreover, the model is capable of learning from small training sets of as few as 10-20 examples.
Future work includes investigation into related biomedical image datasets for the interplay of image synthesis, segmentation, and domain adaptation tasks.

\section*{Acknowledgment}
HZ is supported by the CSC Chinese Government Scholarship.
The project is partially supported by A*STAR JCO grants.
We thank Xiaowei Zhang, Joe Wu, and Malyatha Shridharan for their help with this project.

{\tiny
\bibliographystyle{IEEEtran}
\bibliography{filamentSyn_TMI17}
}

\end{document}